\documentclass[letterpaper]{article} 
\usepackage[]{aaai23}  
\usepackage{times}  
\usepackage{helvet}  
\usepackage{courier}  
\usepackage[hyphens]{url}  
\usepackage{graphicx} 
\usepackage{natbib}  
\usepackage{caption} 
\usepackage{algpseudocode}
\usepackage{algorithm}

\usepackage{newfloat}
\usepackage{listings}
\DeclareCaptionStyle{ruled}{labelfont=normalfont,labelsep=colon,strut=off} 
\floatstyle{ruled}
\newfloat{listing}{tb}{lst}{}
\floatname{listing}{Listing}

\copyrighttext{%
  Distribution Statement A.  Approved for public release. Distribution is unlimited.\\
}

\title{%
  Data-Driven Machine Learning Models for a Multi-Objective Flapping Fin \\
  Unmanned Underwater Vehicle Control System \\\vspace{0.5em}
  \Large{Preprint}%
  }
\author {
    Julian Lee\textsuperscript{\rm 1}\textsuperscript{*},
    Kamal Viswanath\textsuperscript{\rm 2},
    Jason Geder\textsuperscript{\rm 2},
    Alisha Sharma\textsuperscript{\rm 2,3},
    Marius Pruessner\textsuperscript{\rm 2},
    Brian Zhou\textsuperscript{\rm 4}
}
\affiliations {
    \textsuperscript{\rm 1} Yale University, New Haven, CT\\
    \textsuperscript{\rm 2} Naval Research Laboratory, Washington, DC\\
    \textsuperscript{\rm 3} University of Maryland College Park, College Park, MD\\
    \textsuperscript{\rm 4} Thomas Jefferson High School for Science and Technology, Alexandria, VA\\
    \textsuperscript{*}julian.lee@yale.edu\\
    \vspace*{1em}
}

\begin{document}

\maketitle

\begin{abstract}

Flapping-fin unmanned underwater vehicle (UUV) propulsion systems provide high maneuverability for naval tasks such as surveillance and terrain exploration. Recent work has explored the use of time-series neural network surrogate models to predict thrust from vehicle design and fin kinematics. We develop a search-based inverse model that leverages a kinematics-to-thrust neural network model for control system design. Our inverse model finds a set of fin kinematics with the multi-objective goal of reaching a target thrust and creating a smooth kinematic transition between flapping cycles. We demonstrate how a control system integrating this inverse model can make online, cycle-to-cycle adjustments to prioritize different system objectives.

\end{abstract}

\section{Introduction}

Unmanned Underwater Vehicles (UUVs) have the capability to perform various operations including surveillance, exploration, and object detection in underwater environments. While traditional propeller-based propulsion enables UUVs to conduct deep water diving and high-speed traversal, flapping fin propulsion--inspired by the high levels of controllability observed in aquatic animals--offers high maneuverability at low speeds \cite{Blake1979}. Flapping fin propulsion provides a solution for UUVs to effectively navigate near-shore, obstacle-ridden terrain.

Flapping fin UUV control systems regulate vehicle propulsion by modifying the vehicle gait, the specific set of fin kinematics applied to the current flapping cycle. UUV control systems have been extensively explored \cite{He2020}; however, there is sparse literature on flapping fin control. While the effect of various kinematics on propulsion has been studied through experimental \cite{Santo2017}, computational fluid dynamics \cite{Liu2017}, and surrogate model \cite{Viswanath2019} approaches, prior UUV flapping fin control systems do not embed a full understanding of how gait affects propulsion. For example, they focus on experimentally determining a small set of high-propulsion gaits \cite{Shan2019}, restrict chosen gaits to a line in the kinematic space \cite{Bi2014}, or incrementally change kinematics that have a known positive or negative correlation with thrust to eventually reach the target propulsion \cite{Palmisano2008}. We use a neural network model to embed a more comprehensive understanding of the relationship between gait and propulsion within a gait selection model; as a result, the control system can generate gaits that not only meet a target trajectory, but also optimize for other measures of performance such as a smooth transition between gaits and energy efficiency.

\begin{figure}[t]
  \centering
	\includegraphics[width=1.0\linewidth]{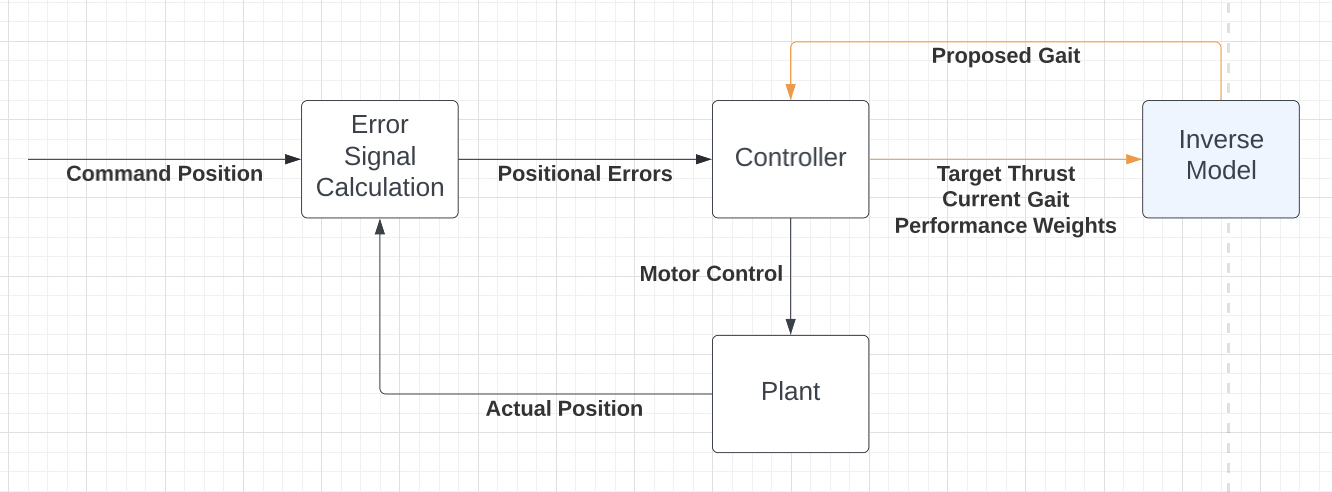}
	\caption{
    Integration of the inverse model (shown in light blue) within a control system for position. Orange arrows indicate one request per thrust cycle.
	}
	\label{fig:config}
\end{figure}

We propose gait generation using a search-based inverse model that invokes a forward surrogate model. Our work focuses on a control system for thrust, which is the forward propulsion of the vehicle. The inverse model determines the subsequent gait from the desired thrust, current gait, and relative performance metric weights for thrust accuracy and kinematic smoothness. Inverse model gait evaluation uses a neural network forward model to predict the thrust of a given gait. Figure \ref{fig:config} shows the integration of the inverse model within the control system.

Search-based methods are frequently used for offline inverse problems as they require invoking the forward model multiple times \cite{Zhou2012,Hansen2017}. Unlike other aerial and underwater vehicle control systems, our flapping fin control system only needs to generate one gait per flapping cycle, allowing for a more flexible time constraint. Therefore, search-based methods are a viable onboard approach that additionally allows for the incorporation of modifiable optimization parameters without retraining the forward model.

As an alternative to a search-based methods, autonomous aerial vehicle literature often uses neural networks in control systems to develop direct inverse models \cite{Muliadi2018,ElHamidi2020}. In this approach, the inverse model consists of a neural network that is directly trained from a forward model or plant. While producing fast predictions, this method does not allow for a flexible optimizer that can be changed cycle-to-cycle by the controller to prioritize different performance metrics.

We demonstrate that our forward gait-to-thrust model accurately interpolates gaits, and we show that our thrust-to-gait search-based inverse model generates high thrust accuracy gaits while embedding adjustable performance weights. These weights allow a control system to make cycle-to-cycle trade-offs between performance metrics based on current system objectives. We compare the performance of different sampling and search-based techniques to improve upon our inverse model performance. Through inverse model integration on the Raspberry Pi, we demonstrate that the inverse model fulfills the one prediction per cycle time constraints, even with the use of an expensive time-series forward model.

\section{Methods}

Our inverse model uses the input from the controller---desired thrust, current gait, and desired performance metric weights---to output a new gait with the goal of minimizing the inverse model loss function. Proposed gaits are generated using a sampling or direct search method. These gaits are then evaluated using a multi-objective loss function that evokes the forward gait-to-thrust neural network model.

\subsection{Loss Function}

Proposed gaits from our inverse model are evaluated with a loss function that is defined as the weighted sum of the loss functions for the different performance metrics. Equation \ref{eq:loss} describes the overall loss function. The losses $L_{t}$, $L_{k}$, and $L_{e}$ correspond to the thrust accuracy, kinematic smoothness, and efficiency loss functions, while $w_{t}$, $w_{k}$, and $w_{e}$ serve as the corresponding performance metric weights.
\begin{equation}
\fontsize{10}{0} L = w_{t} * L_{t} + w_{k} * L_{k} + w_{e} * L_{e}
\label{eq:loss}
\end{equation}

The thrust accuracy loss is computed as the difference between $T_{target}$, the desired thrust, and $T_{pred}$, the thrust from the proposed gait of the inverse model (Eq. \ref{eq:loss_t}). $T_{pred}$ is calculated using the gait-to-thrust neural network.
\begin{equation}
  \fontsize{10}{0} L_{t} = \left|T_{target} - T_{pred}\right|
  \label{eq:loss_t}
\end{equation}

The kinematic smoothness loss accounts for the detrimental effect of frequently moving between gaits with highly deviant kinematics between flapping cycles. Transitioning between similar gaits allows the UUV to undergo a smoother motion, promoting system stability. We define a user-selected equivalent step size such that a change of $s_i$ units for kinematic $i$ has the same kinematic smoothness loss to the system as a change of $s_j$ units for kinematic $j$. The kinematic space is normalized by scaling each dimension based on its equivalent step size; then, kinematic loss is calculated as the Euclidean distance between the current and proposed gait. Equation \ref{eq:loss_k} defines the kinematic smoothness loss function where $n_{k}$ is the number of kinematics, and $x_{i}$ and $y_{i}$ are the values of kinematic $i$ for the current and proposed gait.
  \begin{equation}
    \fontsize{10}{0} L_{k} = \left(\sum_{i=1}^{n_{k}} \left(\frac{|y_{i} - x_{i}|}{s_{i}}\right)^2\right)^{\frac{1}{2}}
    \label{eq:loss_k}
  \end{equation}

Our efficiency loss will be based upon the propulsive efficiency of the gait, which is equal to the product of output thrust and current velocity divided by power. As we do not currently have experimental positive flow cases available for forward gait-to-thrust training, our results set the efficiency weight to 0 and evaluate the trade-offs between thrust accuracy and kinematic smoothness.

\subsection{The Forward Model}

The forward model predicts average UUV thrust for a flapping cycle from the gait. Reduced-order analytic models can produce fast predictions, but they struggle to maintain accuracy when generalized beyond a small parameter space \cite{Muscutt2017}. A model supporting a higher-order input space will allow future forward gait-to-propulsion models to incorporate fluid dynamics-related parameters such as flow speed as well as multi-fin kinematic parameters such as the flapping phase offset between front and rear fins.

\begin{figure}[t]
  \centering
	\includegraphics[width=0.65\linewidth]{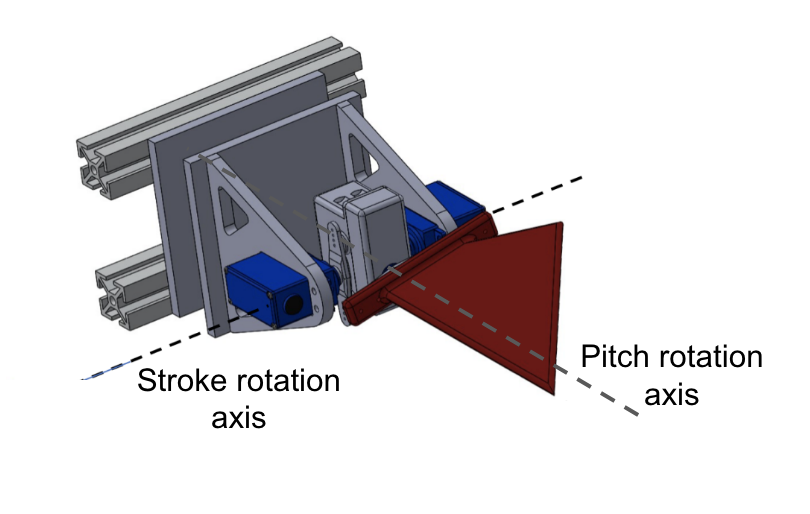}
	\caption{
    Example single pectoral fin setup.
	}
	\label{fig:single_fin_setup}
\end{figure}

Neural network surrogate models support higher degree input spaces, and prior flapping fin propulsion research on fin design has developed neural network surrogate models for thrust prediction \cite{Viswanath2019,Lee2021}. Both works demonstrate that time series models can predict the time history of thrust generation for a flapping cycle. Therefore, we implement a similar long-term short memory (LSTM) network for our gait-to-thrust forward model using the inputs in Table \ref{tab:parameters}. A single pectoral fin setup is shown in Figure \ref{fig:single_fin_setup}. LSTM networks process sequential data by generating an output at each time step and using information from past outputs to inform subsequent results. Compared to traditional recurrent neural networks, LSTM networks include a cell state to retain a long-term memory accumulated from multiple past time steps that influences the next output.

Our control system requests one gait per flapping cycle, so only the average thrust over a cycle is used by our system. While the LSTM model computes average thrust as the mean of the output thrusts, we also test the viability of a simple dense neural network (DNN) model that uses the static kinematic values in Table \ref{tab:parameters} to directly predict average thrust. DNN networks consist of layers of nodes such that each node in layer $l$ is connected to every node in layer $l-1$. Both models are discussed in our results.

\begin{table}[ht]
\footnotesize
\begin{center}
 \caption{Parameters that compose a UUV gait}
 \label{tab:parameters}
 \begin{tabular}{p{3.1cm} p{4.5cm}}
 \hline
 \hline
 Kinematic & Description \\
 \hline
 \hline
 \textit{Static kinematics} &  \\
 \hline
 Stroke Amplitude (°) & Maximum gait stroke angle \\
 \hline
 Pitch Amplitude (°) & Maximum gait pitch angle \\
 \hline
 Flap Frequency (Hz) & Frequency of a stroke cycle \\
 \hline
 Stroke-Pitch Offset & Phase offset of the pitch cycle relative to the stroke cycle, calculated as a fraction of one cycle  \\
 \hline
 \textit{Time-varying kinematics} &  \\
 \hline
 Stroke Angle (°) & Flapping angle as a function of time \\
 \hline
 Pitch Angle (°)  & Pitching angle as a function of time \\
\end{tabular}
\end{center}
\end{table}%

\subsection{Sampling and Direct Search Methods}

Monte Carlo sampling and direct search methods are common derivative-free approaches for optimization based on an objective function \cite{Kroese2014,Audet2014}. Monte-Carlo sampling involves randomly sampling the domain of the objective function based on a probability distribution, while direct search methods use a heuristic approach where past searches influence future attempts. A popular and effective class of direct search methods is the generalized pattern search (GPS) family of algorithms defined by Torczon \cite{Torczon1997}. When applied to optimization problems, GPS often provides fast convergence and high accuracy solutions \cite{Herrera2015,Javed2016}.

For our flapping fin inverse model, each static kinematic listed in Table \ref{tab:parameters} serves as an input for the objective function, and the loss function in Equation \ref{eq:loss} acts as the objective function output. We restrict our input space to the domain of provided experimental data as well as the approximate set of attainable gaits as described in the Experimental Data section. The input space is normalized through compressions in the direction of each static kinematic setting $i$ by the corresponding equivalent step size of $s_{i}$ units; in the normalized input space, any movement between two points of distance $d$ has the same kinematic loss. Both GPS-based and Monte Carlo-based inverse models are implemented as specified below.

\subsubsection{Monte Carlo Approach}

The Monte Carlo method evaluates points from a hyper-sphere centered around the current gait in the input space, where the hyper-sphere radius $a$ determines the search scale. The search scale is calculated as $a = d_t/10$, where $d_t$ is the difference between the thrust of the current gait and the target thrust. A uniform random sample of $n$ points satisfying the inequality in Equation \ref{eq:MCIneq} are evaluated, and the gait with the lowest loss is selected. The variables $x_{i}$ and $y_{i}$ are the value of static kinematic $i$ for the current and new gaits respectively.
\begin{equation}
\fontsize{10}{0}
\sum_{n=1}^{n_{k}} \left(\frac{|y_{i} - x_{i}|}{a}\right)^2 < 1
\label{eq:MCIneq}
\end{equation}

\subsubsection{Generalized Pattern Search Approaches}

Generalized pattern search algorithms provide a fast, derivative-free method for optimization. The objective function is evaluated at points on a mesh generated from a positive spanning set. GPS consists of a search and poll step. During the search step, a finite number of mesh points are evaluated from the objective function using a user-defined procedure; if an improvement is found for any proposed point, then the solution is accepted. If no new point is found, GPS polls neighboring mesh points to the current solution $x_i$ and accepts new points with improved objective function values. If both steps are unable to generate an improved solution, the mesh size $m$ is divided by a mesh size divider $d_m$ and the process is repeated until a certain precision $p$ is obtained. An outline of GPS is provided.

\begin{algorithm}[tb]
\caption{Outline of GPS}
\begin{algorithmic}
    \State Set initial solution $x_{prev}$, initial mesh size $m$, precision $p$;
    \While{$k \ge p$}
         \State $x_{new} \gets x_{prev}$;
         \State SEARCH for new solution in mesh, update $x_{new}$;
         \If{$x_{new}$ == $x_{prev}$}
             \State POLL for new solution in mesh, update $x_{new}$;
             \If{$x_{new}$ == $x_{prev}$}
                  \State $m /= d_m$;
             \EndIf
         \EndIf
    \EndWhile
\end{algorithmic}
\end{algorithm}

Hooke-Jeeves pattern search (HJPS) is a commonly used pattern search algorithm \cite{Hooke1961}. The mesh is created from the positive spanning set consisting of the standard basis vectors. For each input dimension, the poll step evaluates a neighboring mesh point in both the positive and negative directions, and $x_{prev}$ is updated if an improved solution is found. From this step, the vector $x_{new} - x_{prev}$ offers a promising direction for a continued search. The search step leverages this tactic by attempting to move the current solution in the direction $x_{new} - x_{prev}$ until no further improvement can be made.

\begin{algorithm}[tb]
\caption{Search Upon Failure for Our GPS Algorithm}
\begin{algorithmic}
    \State Set current gait and loss as $\vec{g}_{curr}, l_{curr}$
    \State Set direction $\vec{u}_k$ and magnitude $i_k$ of minimum loss increase for each kinematic $k$ during polling
    \State Sort $\vec{u}_k$ by $i_k$ in ascending order
    \State $\vec{v} \gets \sum_{c=0}^{n_k-1} \vec{u}_k$
    \State $j \gets n_k - 1$
    \While{$j \ge 1$}
        \While{$(l_{pred} \gets loss(\vec{g}_{curr} + \vec{v})) < l_{curr}$}
            \State $\vec{g}_{curr} \gets \vec{g}_{curr} + \vec{v}$
            \State $l_{curr} \gets l_{pred}$
            \State $j \gets 0$
        \EndWhile
        \State $\vec{v} \gets \vec{v} - \vec{u}_j$
        \State $j \gets j - 1$
    \EndWhile
\end{algorithmic}
\end{algorithm}

In cases with a high kinematic smoothness weight, movement in any one coordinate direction results in a larger increase to kinematic loss than reduction in thrust accuracy loss, leading to a failed poll step and a resulting failed search step by HJPS. In these situations, HJPS is unable to escape the local minimum until the kinematic smoothness weight is lowered. With the goal of escaping these minima, our proposed generalized pattern search algorithm modifies HJPS to strategically search upon polling failure. Pseudocode for this additional search step is provided. Our GPS algorithm records the direction $\vec{u}_k$ and magnitude $i_k$ of minimum loss increase for each kinematic $k$ during the poll step. Until failure, searches are conducted in the direction of vector $\vec{v}$, where $\vec{v}$ is equal to the sum of all $\vec{u}_k$ and spans all $n_k$ kinematic directions. If the search fails, the component $\vec{u}_k$ associated with the highest loss increase $i_k$ is removed from vector $v$; this process repeats until a search succeeds, or until the vector $v$ only spans one dimension.

\subsection{Experimental Data}

To train and evaluate our forward model, experimental thrust data for various gaits was collected for a setup consisting of rigid rectangular-shaped pectoral fins, one on each side of the UUV. Experiments were run for 864 gaits, which are combinations of the kinematics listed in Table \ref{tab:expData}. Time-series data for five stroke cycles was collected for each gait.

\begin{table}[ht]
\footnotesize
\begin{center}
 \caption{Provided experimental gaits}
 \label{tab:expData}
 \begin{tabular}{p{3.2cm} p{4.1cm}}
 \hline
 \hline
 Kinematic & Provided Values \\
 \hline
 \hline
 Stroke Amplitude (°) & 0, 15, 25, 32, 40, 55 \\
 \hline
 Pitch Amplitude (°) & 0, 15, 25, 32, 38, 55 \\
 \hline
 Flap Frequency (Hz) & 0.75, 1, 1.25, 1.5, 1.75, 2 \\
 \hline
 Stroke-Pitch Offset & -0.0625, 0, 0.0625, 0.125  \\
\end{tabular}
\end{center}
\end{table}%

\begin{figure}[t]
  \centering
	\includegraphics[width=0.6\linewidth]{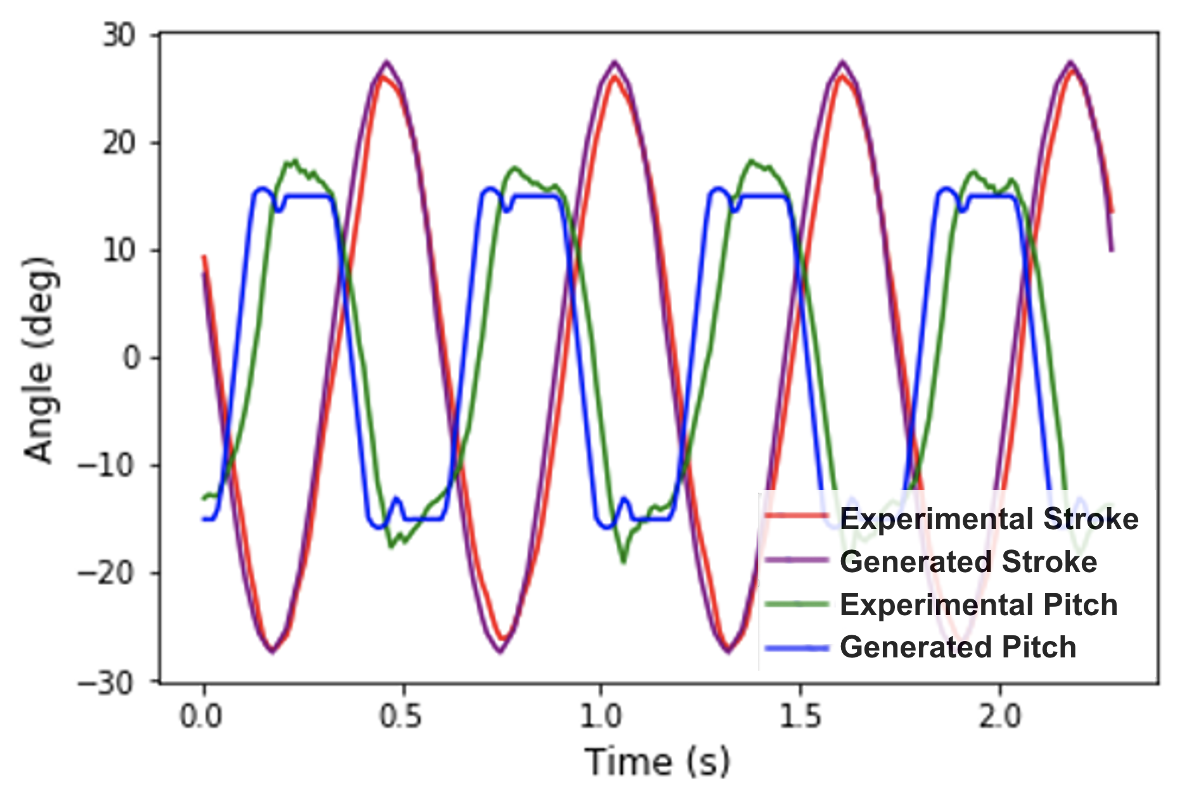}
	\caption{
    Sample stroke and pitch time history.
	}
	\label{fig:strokePitchTimeHist}
\end{figure}

In order to invoke the LSTM network on interpolated gaits, the expected motor stroke and pitch angle time histories must be generated as an input for the network. The motors are commanded 16 values per cycle, where the commanded stroke follows a sinusoidal curve and commanded pitch oscillates between $-p_a$ and $p_a$, where $p_a$ is the pitch amplitude. Time histories are generated using a simple model of motor dynamics accounting for maximum velocity and acceleration. Across the provided experimental data with attainable gaits, mean stroke and pitch angle difference between the generated and experimental time histories was 2.94° and 4.02° for stroke and pitch respectively. An example generated and experimental time history is shown in Figure \ref{fig:strokePitchTimeHist}.

Due to physical system limitations, certain high commanded amplitudes are unattainable at high flap frequencies, $ff$, as the motor is unable to realize the full commanded amplitude within the provided time frame. Therefore, our inverse model excludes all gaits where the stroke amplitude exceeds $97 - ff * 30$ or the pitch amplitude exceeds $75 - ff * 26$. The above inequalities were experimentally determined.

\section{Results and Discussion}

\subsection{Forward Model Performance}

Both the DNN and LSTM were evaluated using the experimental data described in Table 2. Forward passes of the models during evaluation are run using the Tensorflow Lite library \cite{Tensorflow} to increase prediction speed. Z-score normalization was applied to each kinematic input. The DNN directly outputs average thrust over one flapping cycle while the LSTM outputs the full thrust time history for a cycle. The input stroke and pitch time histories for LSTM training and evaluation were generated based on the motor dynamics as described in the Experimental Data section. The LSTM contains 100 hidden units, and time histories consisted of 50 points that were evenly spaced over a flapping cycle. The DNN contains 3 layers with 100 nodes per layer. The LSTM was trained for 150 epochs, and the DNN was trained for 500 epochs.

When trained and evaluated on all experimental gaits, the LSTM and DNN reached mean average thrust errors of 0.0174N and 0.0364N respectively, where the average thrust error refers to the difference between the predicted and experimental average thrust for a specific gait. The DNN produces a significantly higher error, signifying it is unable to effectively learn the gait-thrust relationship across the full space of experimental gaits. The increase in error is concentrated in the high flapping frequency, high amplitude gaits: physically unattainable gaits described in our Experimental Data section have a mean average thrust error of 0.0915 for the DNN. The inability of the DNN to memorize the full gait-thrust relationship poses a concern since the modeling task will grow more complex in the future through the addition of inputs accounting for flow speed and multi-fin kinematic interactions. Therefore, our inverse model will implement the LSTM model.

To test LSTM gait interpolation, a holdout set of gaits was excluded from training. Our holdout set consisted of all experimental gaits fulfilling one or more of the following criteria: a flap frequency of 1.25 Hz, a stroke-pitch offset of 0, or a stroke or pitch amplitude of 25°. The LSTM successfully interpolated kinematics for the excluded gaits with a mean average thrust error of 0.0344N. The worst performing subset of excluded gaits--gaits with a stroke pitch offset of 0--still obtained a mean average thrust error of 0.0374N.

\begin{figure}[t]
  \centering
	\includegraphics[width=1\linewidth]{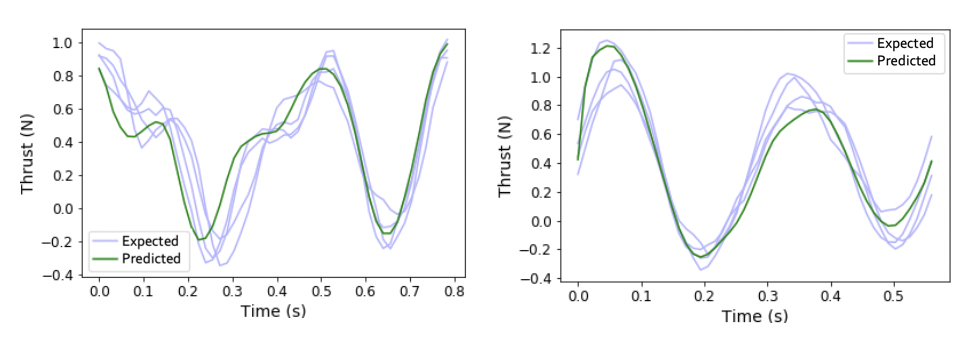}
	\caption{
        Sample thrust time histories for interpolated kinematics. The left graph involves interpolation to an unseen stroke and pitch angle, while the right graph involves interpolation to an unseen flap frequency and stroke pitch offset.
	}
	\label{fig:LSTMThrustProf}
\end{figure}

Figure \ref{fig:LSTMThrustProf} shows example thrust time histories generated by the LSTM for interpolated gaits. The LSTM embeds an understanding of how thrust changes over the course of a flapping cycle, capturing the peak and troughs of the thrust time history; this understanding offers an explanation for the high-accuracy LSTM average thrust predictions for interpolated kinematics.

\subsection{Inverse Model Performance}

Three search-based methods for the inverse model were evaluated--Monte Carlo, Hooke-Jeeves Pattern Search (HJPS), and our Generalized Pattern Search (GPS) algorithm that builds upon HJPS. Monte Carlo generates 50 trial gaits ($n = 50$). The mesh size $k$ and precision $p$ for both pattern search algorithms is set to 3 and 0.375 respectively. The inverse models implement the LSTM forward model for gait-to-thrust prediction.

Synthetic and simulated thrust requests were used to evaluate the inverse models. Each synthetic data set consists of a sequential list of 100 pseudo randomly generated thrust requests with a difference between 0 and $\Delta T_{max}$ for adjacent thrusts. Thrust requests were restricted to the range 0.2N to 1.2N, and the value of $\Delta T_{max}$ was varied from 0.1N to 1.0N in increments of 0.1N to produce 10 data sets.

\begin{figure}[t]
  \centering
	\includegraphics[width=1\linewidth]{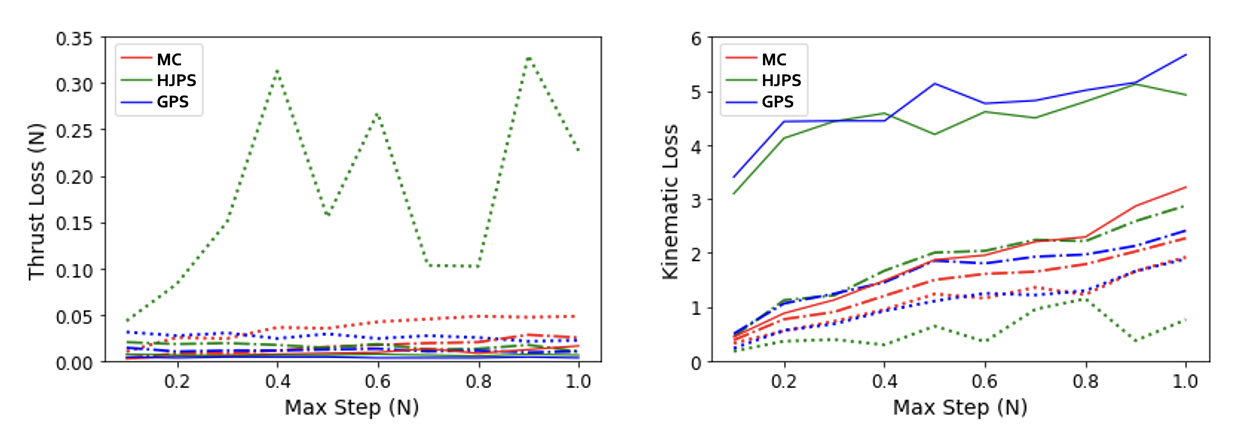}
	\caption{
        Inverse model performance for synthetic data, measured by thrust accuracy loss (left) and kinematic smoothness loss (right). The dotted, dashed, and solid lines have thrust weights of 0.9, 0.95, and 1. Kinematic weights are set to $1 - w_{t}$.
	}
	\label{fig:SynDataInvModelResults}
\end{figure}

Inverse model performance on synthetically generated thrust requests is shown in Figure \ref{fig:SynDataInvModelResults}. As the maximum step size increases, the average kinematic loss generally trends upwards for all three models: the inverse models can obtain gaits closer in the kinematic input space for smaller changes in thrust. An increase in thrust weight for our inverse models results in a lower thrust loss and a higher kinematic loss. Therefore, priorities on different objectives--thrust accuracy and kinematic smoothness--can be directly regulated by a control system through a change in weights.

The models do not optimize for kinematic smoothness in the cases where the models have a thrust weight of 1, yet Monte Carlo still exhibits a low kinematic smoothness loss across the tested input thrust data. Compared to HJPS and GPS, the MC model has a tendency to find closer gaits in the input space without implicitly embedding kinematic smoothness into the loss function. However, HJPS and GPS successfully obtain similar kinematic smoothness values to MC for higher kinematic smoothness weights.

The GPS and MC inverse models demonstrate a high thrust accuracy performance across all three weights, achieving an average thrust loss of less than 0.04N. While the HJPS inverse model performs similarly to the GPS model for higher thrust weights, this model shows a significant deterioration in thrust accuracy when a higher kinematic smoothness weight of 0.1 is applied. The higher kinematic smoothness weight increases the overall cost of moving to a new gait; in these cases, the HJPS inverse model reaches a gait where movement in any coordinate direction results in a higher increase in kinematic smoothness loss than reduction in thrust accuracy loss. At this point, the algorithm becomes trapped at a local minimum for all subsequent iterations unless the thrust accuracy weight is increased. Our GPS method strategically searches promising points in situations where the algorithm is potentially trapped in a local minimum, namely during polling failure. These additional searches enable the GPS inverse model to maintain a low thrust loss for lower thrust weights.

\begin{figure}[ht]
  \centering
	\includegraphics[width=1.0\linewidth]{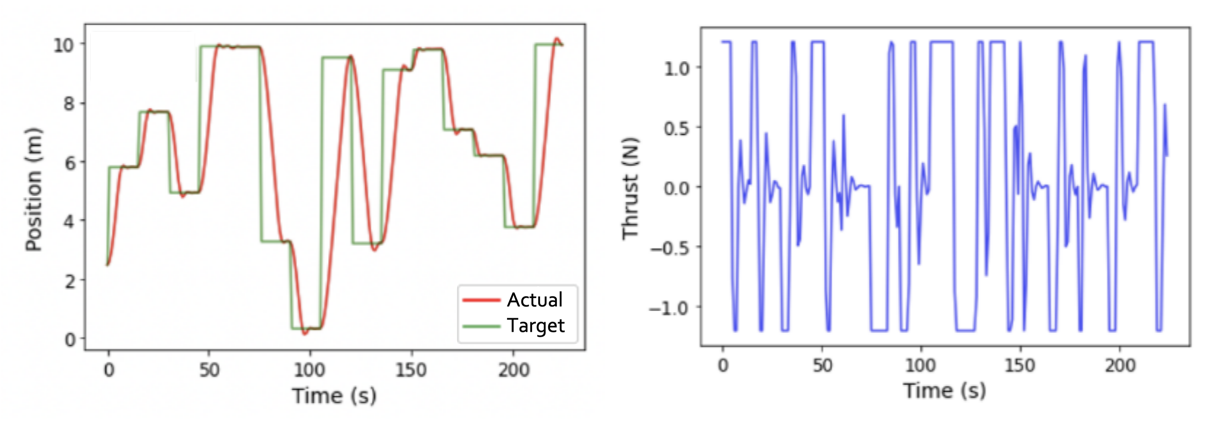}
	\caption{
     Sample position (left) and thrust request (right) time histories from the simulated PID controller for position.
	}
	\label{fig:PID_Pos_Thrust}
\end{figure}

To evaluate our inverse models on more realistic thrust requests, a PID controller for vehicle position was used for thrust request generation. The PID controller outputs a target thrust to control a simulated thrust-to-position plant that models the UUV based on the translational equations of motion for a rigid body. We simulate UUV movement to 100 randomly generated positions between 0 and 10m. For each location, the controller commands a thrust between -1.2N and 1.2N for 15 flapping cycles to reach and then remain at the position. A sample of the change in positional values and corresponding thrust requests from the simulation is provided in Figure \ref{fig:PID_Pos_Thrust}. The inverse model generated kinematics is able to accurately track the target trajectory in the given timeframe. This experiment uses small positional changes to simulate the start and stop conditions of the UUV: for intermediary movement, the UUV thrust is held constant at the maximum or minimum value.

As provided experimental data does not cover the domain of negative-thrust gaits producing negative trusts, the inverse model temporarily assumes symmetry in the gait landscape. A gait with a negative stroke pitch offset, stroke amplitude, and pitch amplitude generates a thrust $-T$, where its positive counterpart generates a thrust $T$. For this symmetry assumption to hold true, it would physically require a fin rotation of 180° along the pitch axis during every transition between a positive and negative gait such that the leading edge of the fin faces the direction of movement. This temporary solution allows for the inverse model to access negative-thrust gaits for simulated data testing, and the kinematic landscape will be extended to incorporate negative-thrust gaits in the future.

\begin{figure}[t]
  \centering
	\includegraphics[width=1\linewidth]{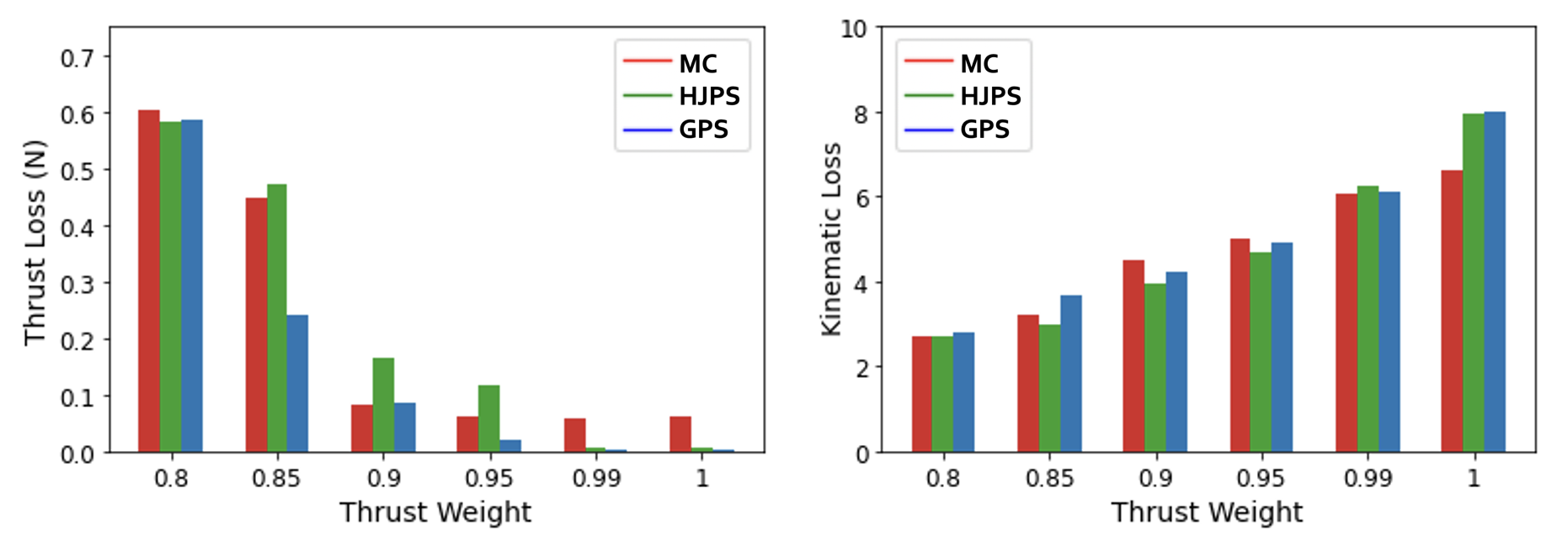}
	\caption{
        Inverse model performance for simulated data, measured by thrust accuracy loss (left) and kinematic smoothness loss (right). The kinematic smoothness weight was set to $1-w_t$.
	}
	\label{fig:SimDataInvModelResults}
\end{figure}

The inverse models are evaluated on the 1500 thrust requests generated by the simulation. To emulate the onboard system, inverse models were run on a Raspberry Pi 4 Model B. Model performance across various weight settings is shown in Figure \ref{fig:SimDataInvModelResults}. Table \ref{tab:simInvModelResults} summarizes the performance of MC, HJPS, and GPS across weight settings: losses are averaged across weight settings, and the worst-case run time across all calls to each algorithm is provided.

\begin{table}[ht]
\footnotesize
\begin{center}
 \caption{Inverse Model Simulation Performance Summary}
 \label{tab:simInvModelResults}
 \begin{tabular}{p{2.8 cm} p{1.2cm} p{1.2cm} p{1.2cm}}
 \hline
 \hline
  & MC & HJPS & GPS \\
 \hline
 \hline
 Thrust Loss (N) & 0.219 & 0.224 & 0.157 \\
 \hline
 Kinematic Loss & 4.693 & 4.747 & 4.940 \\
 \hline
 Overall Loss & 0.484 & 0.469 & 0.435 \\
 \hline
 Maximum Time (s) & 0.300 & 0.393 & 0.467  \\
\end{tabular}
\end{center}
\end{table}%

Onboard predictions must meet the time constraints of 1 generated gait per flapping cycle. As the onboard UUV will not exceed a flapping frequency of 2 Hz, the inverse model must generate a gait within 0.5 seconds. All inverse models consistently run within 0.5 seconds on the Pi: the GPS algorithm has the slowest maximum run time of 0.467 seconds. Therefore, these inverse models are all viable for an onboard flapping fin UUV control system.

All algorithms were responsive to changes to performance metric weights: higher thrust weights resulted in lower thrust losses and higher kinematic losses. GPS reached the lowest mean overall loss of 0.435, as well as a mean thrust accuracy of less than 0.01N when a thrust weight of 1 is applied. While the inverse models demonstrate similar performances when minimizing kinematic loss, GPS consistently outperforms MC and HJPS in terms of minimizing thrust loss as seen in Figure \ref{fig:SimDataInvModelResults}. GPS obtains a mean thrust accuracy loss of 0.157N averaged across the tested weights, while MC and HJPS obtain thrust accuracy losses of 0.219N and 0.224N. The inverse models demonstrate similar performances when minimizing $L_k$, obtaining mean kinematic smoothness losses of 4.693, 4.747, and 4.940 respectively.

\section{Conclusion and Deployment Strategy}

Our work uses neural networks to embed a deep kinematic-thrust relationship in a flapping fin UUV control system with the goal of multi-objective optimization: more specifically, we design a search-based inverse model that invokes a gait-to-thrust forward model to select gaits for the controller.

We demonstrate that our LSTM forward model effectively learned the full space of kinematic-thrust mappings and accurately interpolated to unseen gaits; the DNN model, which does not use time-series data, was unable to match the performance of the LSTM. We implemented inverse models incorporating the LSTM forward model using three approaches: Monte-Carlo Sampling, Hooke-Jeeves Pattern Search, and our new Generalized Pattern Search method, which is an extension of HJPS. Upon integration with onboard hardware, the inverse models consistently ran within the time constraint of 0.5 seconds per iteration. When evaluated with simulated PID controller thrust requests, our GPS algorithm yielded the best performance for minimizing both thrust loss and overall loss across weight settings.

All three inverse models successfully made trade-offs between thrust accuracy and kinematic smoothness based on the applied performance metric weights. Our flexible inverse model framework enables future UUV control systems to incrementally adjust the emphasis placed on different measures of performance based on the current task and vehicle status. For example, the emphasis on thrust accuracy can be dynamically changed by the controller based on the degree of precise maneuvering required for the task at hand. Our inverse model framework also allows for the incorporation of additional performance metrics such as efficiency.

Prior to deployment, we will collect experimental data for gaits producing negative thrusts and subsequently retrain the forward model based on this new data. Since our inverse model is already integrated with onboard hardware, the Raspberry Pi 4, the final deployment step consists of establishing onboard communication between the Raspberry Pi containing our inverse model and a PID-based micro controller; at this point, physical experiments can be conducted using the inverse model. Following deployment, our inverse model will be extended to incorporate the propulsive efficiency performance metric, which is calculated as the product of output thrust and current velocity divided by power; inverse model performance for thrust accuracy and efficiency trade-offs will be evaluated once experimental data for positive flow speeds becomes available.

\bibliography{ms}

\appendix

\end{document}